\documentclass[journal,transmag]{IEEEtran}
\usepackage{textcomp}
\IEEEoverridecommandlockouts
\usepackage[utf8]{inputenc}
\usepackage{cite}
\usepackage{graphicx}

\usepackage{caption}

\begin{document}

\title{Automatic Main Character Recognition for Photographic Studies \thanks{We would like to acknowledge support from Helsingin Sanomat foundation, project "Machine learning based analysis of the photographs of the corona crisis" and from Intelligent Society Platform funded by Academy of Finland, project “Improving Public Accessibility of Large Image Archives”.}}
\author{\IEEEauthorblockN{Mert Seker\IEEEauthorrefmark{1},
Anssi Männistö\IEEEauthorrefmark{2},
Alexandros Iosifidis\IEEEauthorrefmark{3} and
Jenni Raitoharju\IEEEauthorrefmark{4}}

\IEEEauthorblockA{\IEEEauthorrefmark{1}Unit of Computing Sciences, Tampere University, Finland (e-mail: mert.seker@tuni.fi)}
\IEEEauthorblockA{\IEEEauthorrefmark{2}Unit of Communication Sciences, Tampere University, Finland
(e-mail: anssi.mannisto@tuni.fi)}
\IEEEauthorblockA{\IEEEauthorrefmark{3}Department of Eletrical and Computer Engineering, Aarhus University, Denmark (e-mail: ai@ece.au.dk)}
\IEEEauthorblockA{\IEEEauthorrefmark{4}Programme for Environmental Information, Finnish Environment Institute, Jyväskylä, Finland (e-mail: jenni.raitoharju@syke.fi)}% <-this % stops an unwanted space
}

\markboth{Seker \it{et al}.: Automatic Main Character Recognition for Photographic Studies}{Seker \it{et al}.: Automatic Main Character Recognition for Photographic Studies}

\maketitle

\begin{abstract}
Main characters in images are the most important humans that catch the viewer's attention upon first look, and they are emphasized by properties such as size, position, color saturation, and sharpness of focus. Identifying the main character in images plays an important role in traditional photographic studies and media analysis, but the task is performed manually and can be slow and laborious. Furthermore, selection of main characters can be sometimes subjective. In this paper, we analyze the feasibility of solving the main character recognition needed for photographic studies automatically and propose a method for identifying the main characters. The proposed method uses machine learning based human pose estimation along with traditional computer vision approaches for this task. We approach the task as a binary classification problem where each detected human is classified either as a main character or not. To evaluate both the subjectivity of the task and the performance of our method, we collected a dataset of 300 varying images from multiple sources and asked five people, a photographic researcher and four other persons, to annotate the main characters. Our analysis showed a relatively high agreement between different annotators. The proposed method achieved a promising F1 score of 0.83 on the full image set and 0.96 on a subset evaluated as most clear and important cases by the photographic researcher.
\end{abstract}


\section{Introduction}

Recognizing and analysing the main character is traditionally one of the basic tasks in wide variety of photographic studies in the fields of social sciences and humanities \cite{dodd1989face, kdra2018refugees}. This approach is essential in research concerning questions of salience or bias of media content or when analysing historical changes in modes of representation of, for example, gender, occupation, social class, or ethical issues \cite{bell2011}. Studying the contents of photographs in mass media or historical archives has long traditions especially in qualitative research, but also quantitative studies over large amounts of photographs have become more mainstream.

A turning point enhancing the use of quantitative methods was the introduction of a unified model for visual content analysis by Kress and van Leeuwen in 1996 \cite{kress_leeuwen_1996}. Their model has been ever since widely used in several research contexts, and the model itself has been developed further by other scholars \cite{bell2011, jewitt_oyama_2011}. One of the starting points for subsequent analysis in the model is defining the main character and his/her relation to other persons  in the photo. The model provides an approach for disassembling the semiotic structure of images in a way which enables to study the structure, actors, and activities in an image with quantitative tools. As it is described in \cite{bell2011}:
“Visual content analysis is a systematic, observational method used for testing hypotheses about the ways in which the media represent people, events, situations, and so on. It allows quantification of samples of observable content classified into distinct categories. It does not analyse individual images or individual ‘visual texts’.”

In traditional photographic research settings, the amount of photographs is somewhere between 400-1000. As some examples, a systematic inventory of the types, range, and frequency of photographic images presented in American news-magazines during Persian Gulf War with 1104 images was conducted in \cite{griffin1995, griffin2004}. Stereotyped portrayals of males and females in magazine advertisement were analyzed with 827 images in \cite{bell_and_milic_2002}. A study in \cite{mannisto2004} compared the topics of photographs of Afghanistan War (2001, 204 photos) and Iraq War (2003, 343 photos) in Finnish newspapers, and death images in Israeli newspapers were studied with 506 photos in \cite{morse2015}. The relatively low number of images in such studies follows from the laborious manual analysis required for conducting the research. If a study focuses on human behaviour or visibility in photos (as the typical cases are) even selecting – not to mention the actual research process of analysing photographs individually – the photos depicting people is laborious.

Contrary to the traditional methods, the introduction of novel machine learning methods has potential to both significantly reduce the time for selecting the suitable material and to give quantitative and detailed observations of the contents of very large collections. The model provided by Kress and van Leeuwen \cite{kress_leeuwen_1996} is particularly suitable to be used together with machine learning tools, because it defines some basic concepts of photographic analysis  in a manner that enables automatic analysis. For the sake of this paper, it is noticeable how Kress and van Leeuwen define the main character in a photo: “They are often also the most salient participants, through size, place in the composition, contrast against background, colour saturation or conspicuousness, sharpness of focus, and through ‘psychological salience’, which certain participants (…) have for the viewers”. 

In this paper, we propose a method for automatically recognizing the main character(s) in photos. We use machine learning based person detection and consider some of the properties defined by Kress and van Leeuwen to determine whether a detected person is a main character. After recognizing the main character, the analysis can be continued by applying other automated techniques such as gender or age recognition \cite{agegender0,agegender01}, scene recognition \cite{scenerecognition1,scenerecognition2}, interaction \cite{interactionrecognition1,interactionrecognition2} and distances to other characters \cite{KORTE,proxemics}. Overall, the novel machine learning techniques will allow conducting detailed quantitative photographic studies on very large collections of photos and, thus, renew the whole field of photographic studies in the near future.

The rest of this paper is organized as follows. We explain the most similar works to our proposed method in the literature in Section \ref{sec:related}, Section \ref{sec:method} provides the detailed explanation of our proposed method, Section \ref{sec:results} introduces our experimental setup and results. Finally, we conclude the paper in Section~\ref{sec:conclusion}.

 \section{Related Work}
\label{sec:related}

While there have been some efforts to identify important character(s) in photos, both the motivation and definition of the problem vary. Commonly, the important person detection task is motivated as a tool to better understand what is happening in the image, which can be used as a substep in various computer vision tasks such as event recognition or image captioning \cite{VIP, personrank, li2019point}. Some works focus on finding images of specific person as main characters to assist automatic photo album creation \cite{dedeyter2018photoalbums}.

In \cite{personrank}, two publicly available datasets for important person detection were collected by asking ten people to label the single most important person in each photo. Thus, the annotations follow average intuitive opinion on the most important person and the setup ignores the question of how many people in each image should be considered as main characters. In photographic studies, the number of main characters is also an important question and, based on our experience, it is one of the main sources of uncertainty both in annotations and for the algorithm. The work in \cite{VIP} also used a pool of annotators to collect their dataset. They found that there was low inter-human agreement if they asked people to simply annotate the important persons. Therefore, they reformulated the problem into pair-wise importance ranking for every pair of detected people. In GAF-personage dataset \cite{grouplevel}, the most influential person was annotated by three different persons who were first given some instructions based on a survey. For example, the annotators were instructed to favor children in family scenarios.

Methods proposed in the literature use various criteria for deciding the most important person. The method proposed in \cite{VIP} relies on low-level cues based on location, size, sharpness, facial pose, and occlusion. For this purpose, they apply a face detection algorithm and then use traditional static feature extraction techniques to estimate the importance based on the criteria. The method proposed in \cite{personrank} labels the person with the highest estimated activity level as the most important person. To this end, a graph based approach is used, where every person appearing in an image is represented as a node and the interactions between them as edges. The interactions are represented as the union of two separate graphs: a pair-wise bidirectional graph and a unidirectional hyper-interaction graph. This hybrid graph is called Hybrid-Interaction Graph (HIG). Finally, a modified version of the PageRank \cite{pagerank} algorithm called PersonRank is used to estimate the activeness of each person. The study in \cite{li2019point} uses a deep relation network called deep imPOrtance relatIon
NeTwork (POINT) that infers person to person and person to event interactions in images. Then, these two kinds of interaction are used to compute features for selecting the most important person. The work in \cite{grouplevel} explored the use of group affect and emotional intensity in detecting the most influential people. To overcome the difficulty of collecting large datasets with the most important person annotated for training learning-based approaches, a semi-supervised approach was proposed in \cite{hong2020semi}.

\section{Proposed Method}
\label{sec:method}

As we approach the problem of main character recognition with the intention to assist large-scale photographic studies, we use the main character definition from Kress and van Leeuwen~\cite{kress_leeuwen_1996} as it is common in the field. We aim at recognizing all the main characters, not finding the single most important person or ranking people based on their importance as the related works introduced above. To the best of our knowledge, there are no earlier works on automatic main character detection from this perspective. 

Our proposed method uses a deep learning based approach for detecting people and on top of this static feature extractors that match with the Kress and van Leeuwen definition. In this sense, our method has some similarities with \cite{VIP}, but in \cite{VIP} person detection is based on face detection, while we rely on human pose estimation, which we show to be much more robust. We also rely on a different set of static features. As only the person detection part is learning-based, we do not need any main character annotations for training our approach, which is important considering that there are no datasets matching our objectives available. 

We take advantage of the OpenPose \cite{openpose1,openpose2,openpose3,openpose4} human pose estimation model to detect people in photos and define facial rectangles on each detected person. The main advantages of using human pose estimation instead of object detection or face detection are as follows: The facial rectangles can still be drawn even if the face is partially obstructed or if the subject is facing away from the camera as long as at least two head keypoints are detected. The bounding boxes extracted by object detectors tend to overlap when the subjects are close to each other and this is problematic for cropping individual subjects from the image. 

\begin{figure}[!t]
    \centering
    \includegraphics[scale=0.27]{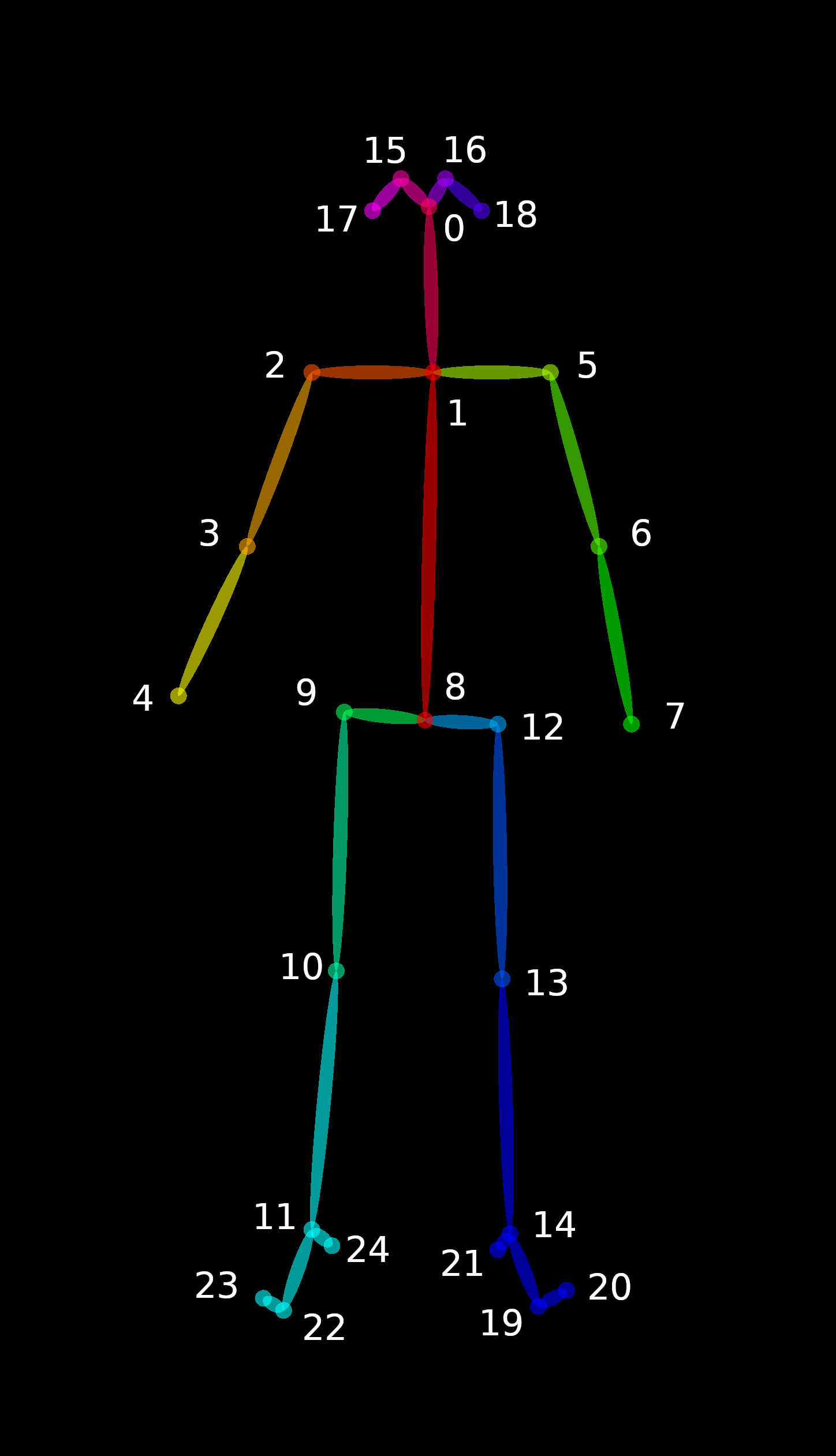}
    \caption{25 skeleton keypoint output of OpenPose \cite{openpose1,openpose2,openpose3,openpose4}.}
    \label{fig:openposekeypoints}
    \vspace{-5px}
\end{figure}
\begin{figure}[!t]
    \centering
    \includegraphics[scale=0.28]{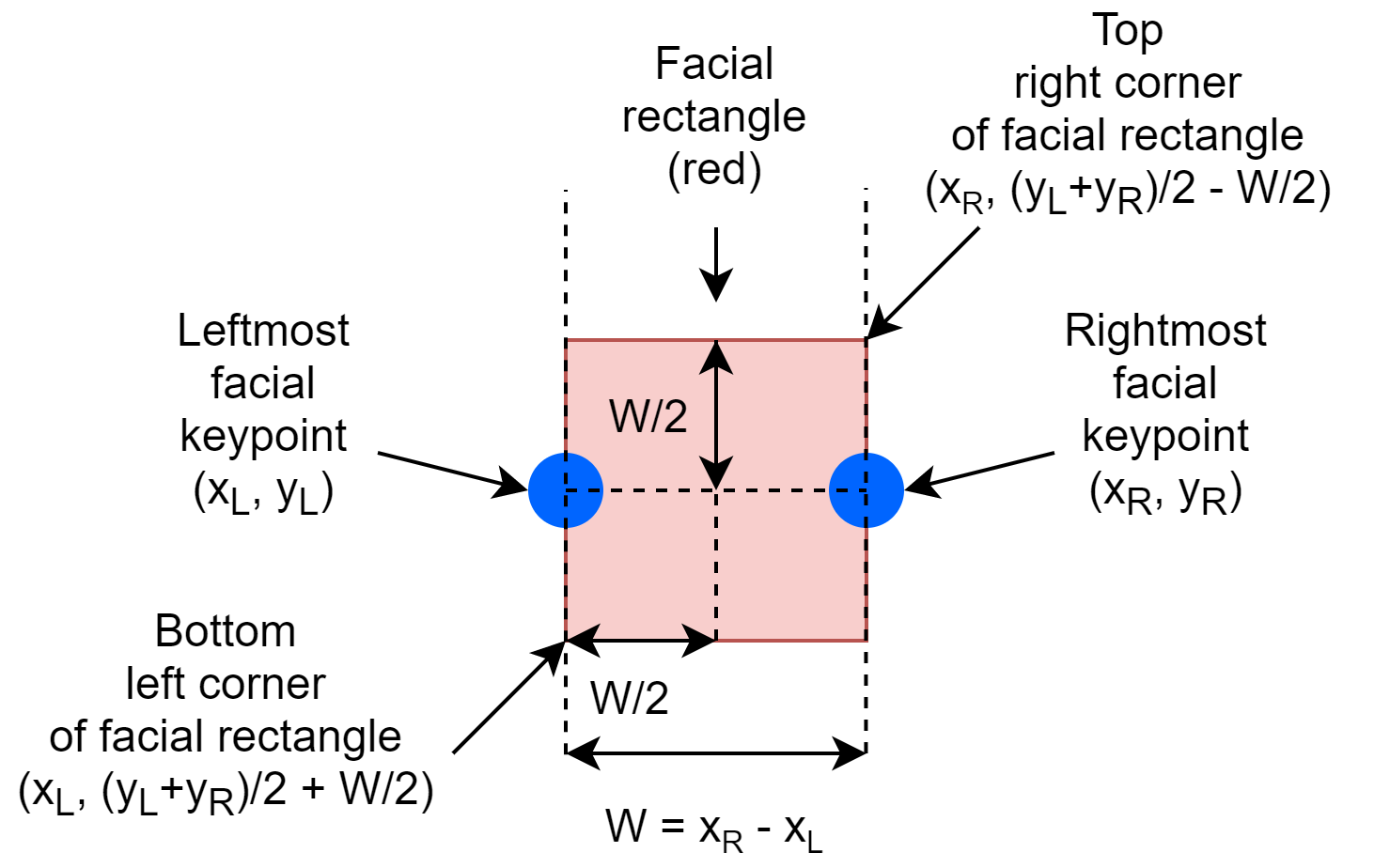}
    \caption{Calculation of the facial rectangle corner positions.}
    \label{fig:rectangleequation}
    \vspace{-10px}
\end{figure}

We use the 25 keypoint output version of OpenPose shown in Fig.~\ref{fig:openposekeypoints} and consider the head keypoints 17, 15, 0, 16, and 18 for cropping the rectangles as follows: the position of the leftmost facial keypoint with the smallest x value is denoted as $(x_{L}, y_{L})$ and the position of the rightmost facial keypoint with the largest x value is denoted as $(x_{R}, y_{R})$. The width of the drawn facial rectangles is $W = x_{R} - x_{L}$. The positions of the bottom left and top right corners of the facial rectangles are given as $(x_{L}, (y_{L}+y_{R})/2 + W/2)$ and $(x_{R}, (y_{L}+y_{R})/2 - W/2)$, respectively. Fig.~\ref{fig:rectangleequation} shows an illustration of the calculation of the facial rectangle corner positions. 
Example images with facial rectangles drawn on the detected persons are given in Fig. \ref{fig:goodexamples}. 

After cropping the facial rectangles from the full image, they are processed individually to generate an overall importance score for each person.  Here, we consider three of the properties defined by Kress and van Leeuwen: size, place in the composition, and sharpness of focus. We also consider gaze direction, which can be seen as a factor of psychological salience mentioned in the definition of Kress and van Leeuwen.

We consider the area of the facial rectangles for the area scores, closeness of the facial rectangles to the center of the photo for the place scores, and edge detection on the cropped facial images for the sharpness scores. The area scores are given according to the pixel-wise area of the facial rectangles: $W*W$. The position scores are given as $D/2 - d$ where $D$ is the pixel-wise diagonal length of the image and $d$ is the pixel-wise direct distance of the center of facial rectangles to the center of the image. The sharpness scores are computed using the Canny edge detection method \cite{canny} with 50 and 250 as the minimum and maximum threshold values on the cropped facial rectangles. Canny edge detection outputs a two dimensional array that has the same pixel size as the cropped facial rectangles. The detected edge pixels have a value of 255 while the rest of the pixels have a value of 0. We compute the mean of this two dimensional output of the Canny edge detection method to generate a sharpness score. Since sharper faces tend to have more edges, applying edge detection works well in most of cases. 

We do not set specific thresholds for these scores due to the variations between the images. Instead, we compute the scores of each person by normalizing the values in each score type for each image: The largest face in a photo will have an area score of 1, the sharpest face will have a sharpness score of 1, the face closest to the center of the image will have a position score of 1, and the scores for other persons are set relative to these. We use the weighted sum of these scores to compute the final importance score each person.

Moreover, we infer the gaze direction of the persons by considering which of the head keypoints were detected by OpenPose. We consider a subject to be facing away from the camera if only the ear keypoints (17 and 18) are visible and the facial keypoints (15, 16 and 0) are not visible. If a subject is detected to be facing away from the camera, their overall score is multiplied by 0 so that they will never be picked as the main character. We also take the average prediction confidence of the head keypoints into account. If the average confidence is lower than 0.45, the subject's overall score is multiplied by 0. Moreover, OpenPose can sometimes falsely connect the predicted body parts of different people as if they belong to the same person. This results in a significantly large area score for the faulty detection which prevents the method from selecting the true main characters. In order to alleviate this problem, we discard the detections where the left (from the viewer's perspective) eye is on the right side of the right eye or where the direct distance between the ears is larger than 50 times the direct distance between the eyes.  

The overall equation for the importance score $S$ is
\begin{equation}
    S = (w_aS_{a} + w_sS_{s} + w_pS_{p}) * G,
    \label{eq:overallscore}
\end{equation}
where $S_{a}$, $S_{s}$, and $S_{p}$ are the area, sharpness, and position scores, respectively, $w_x$ denote the corresponding weights, and $G$ denotes the gaze direction, which take the value of 0 if the person is facing away from the camera and 1 otherwise. 

After the overall scores are calculated for each person, they are normalized again so that the most important person in a photo gets an overall score of 1. Thus, the normalized overall scores give the importance of each character in the range [0,1]. Finally, we compare the scores for all the detected people to decide who is a main character. For each image, we classify at least one person as the main character, but there may be also several main characters. To this end, we experimentally define a threshold $T$  and choose all the persons who have a normalized overall score higher than this threshold as the main characters. 

As our method was developed in collaboration with a photographic researcher, we adjusted our hyperparameters manually to match with his impression of the main characters. This approach was selected instead of data-based approaches (e.g., cross-validation), because for the subsequent use in photographic studies it was essential to consider also the expert's opinion on severity of different errors and not just blindly optimize the overall accuracy. With this approach, we obtained the following hyperparameter values: $w_a = 3.5$, $w_s = 3$, $w_p = 1.2$, and $T = 0.92$ for images with more than two detected people and $T = 0.86$ for images with two detected people. The lower threshold for the two-person images is because we observed that in such images it is more likely that both persons are main characters. We set these hyperparameter values with a diverse set of images, but we also observed that for certain image types it might be beneficial to re-optimize the values. For example, historical images tend to have focusing problems due to the technological limitations of the cameras at the time and, therefore, sharpness is not as reliable indicator as for modern images. 

We make our codes publicly available\footnote{https://github.com/mertseker-dev/main-character-detection}. The method can be easily used on any input image and outputs the facial rectangle locations of the detected main characters. 

\section{Results and Analysis}
\label{sec:results}

\subsection{Test Dataset}
\label{sec:dataset}
To evaluate the performance of our main character detection algorithm, we collected a total of 300 images from different sources in order to have a large variety of images and annotated the main characters. Out of the 300 images, 40 and 60 were collected from the public websites  of Finnish newspapers Aamulehti\footnote{https://www.aamulehti.fi/} and Helsingin Sanomat\footnote{ https://www.hs.fi/},  respectively, 100 were collected from the public website of Finnish Wartime Photograph Archive\footnote{http://sa-kuva.fi/} which contains pictures from the Second World War, 80 were collected from Google OpenImages dataset\footnote{https://storage.googleapis.com/openimages/web/index.html}, and 20 were collected from the KORTE dataset\footnote{https://doi.org/10.23729/b2ea87e6-b845-46b8-abf3-cdbe299ce8b0}\cite{KORTE}. 

The annotations were done by five different people, one of whom is a photographic researcher. He annotated the images following the main character definition of Kress and van Leeuwen as closely as possible. The other four annotators were not aware of the definition and just followed their subjective opinion. All annotators were asked to annotate at least one main character for each image, but there was no upper limit for the amount of main character annotations on a single image. We consider the annotations from the photographic researcher as our ground-truth. He annotated a total of 373 main characters. We also asked him to label images either as clear or unclear cases. He labeled 230 images as clear cases, where any errors can be considered more severe, and 70 as unclear cases, where several solutions could be acceptable following the definition of Kress and van Leeuwen. 

\subsection{Experimental Setup}

As we approach the problem as a binary classification task where each detected person is classified as a main character or not, we use the standard binary classification evaluation metrics: Precision, Recall and F1 score. We consider the main characters as the positive class. We do not explicitly consider the human detection accuracy, but we consider every non-detected person as a negative classification.

We used OpenPose pretrained on the COCO keypoint challenge dataset \cite{COCO} where the body parts of people are annotated as well as another dataset that consists of a subset of the COCO dataset where the foot keypoints are annotated. The hyperparameters were set as explained in Section~\ref{sec:method}. 

We were not able to conduct comparative analysis against important person detection techniques discussed in Section~\ref{sec:related}, because the codes were either not available or not operational without significant changes and the papers did not provide enough details to replicate the methods. Furthermore, we could not compare against the previously published results on public datasets given in \cite{personrank,li2019point}, because the used evaluation metric is not accurately defined. It should be also remembered that the objective of these important person detection techniques is different, so any direct comparisons would be somewhat ambiguous. To justify the use of human pose estimation instead of direct face detection, we applied our method also with a Max-Margin \cite{king2015maxmargin} convolutional neural network (CNN) face detector. The model was trained on a dataset containing 7220 images, collected from various datasets such as ImageNet \cite{ImageNet} and PASCAL VOC \cite{PASCALVOC}. The detected facial rectangles of the face detector were used in the algorithm instead of the facial rectangles generated from the human pose estimation output. This approach has a similar main pipeline as the method proposed in \cite{VIP}.

\subsection{Results and Analysis}
\label{ssec:results}

\begin{figure*}[!t]
    \centering
    \centering
    \includegraphics[width=0.32\linewidth]{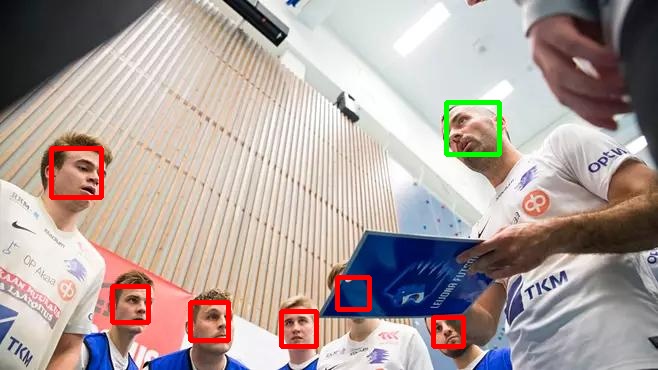}
    \includegraphics[width=0.32\linewidth]{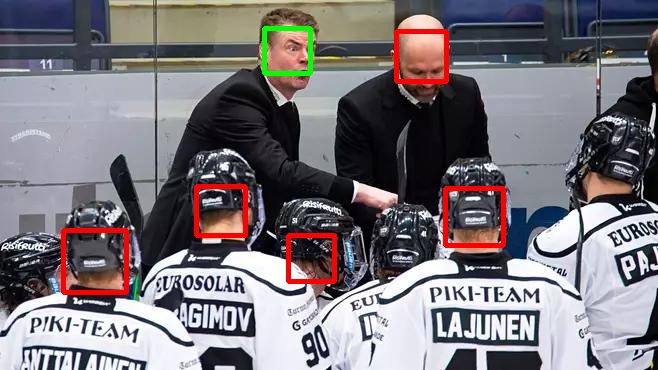}
    \includegraphics[width=0.32\linewidth]{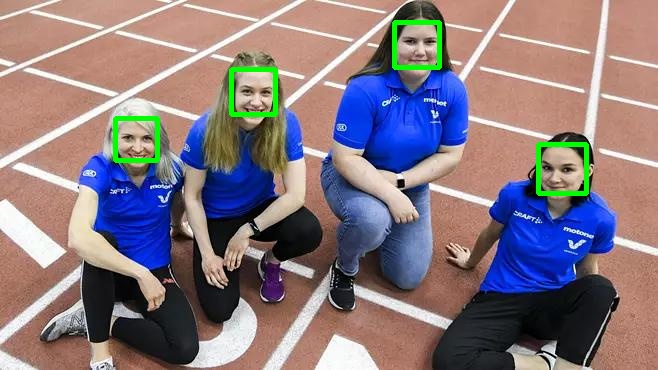} \\
    \vspace{5pt}
    \includegraphics[width=0.32\linewidth]{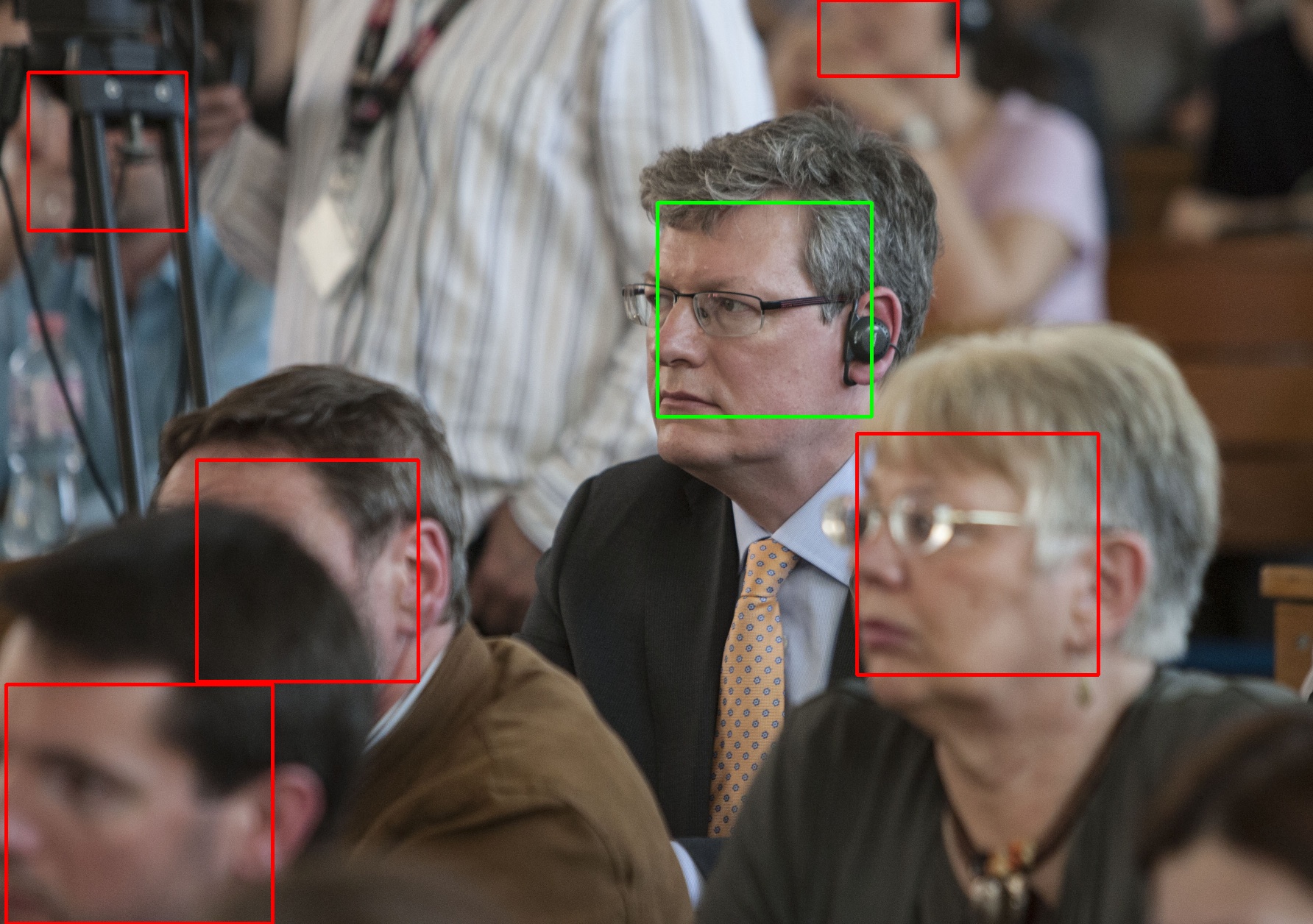}
    \includegraphics[width=0.32\linewidth]{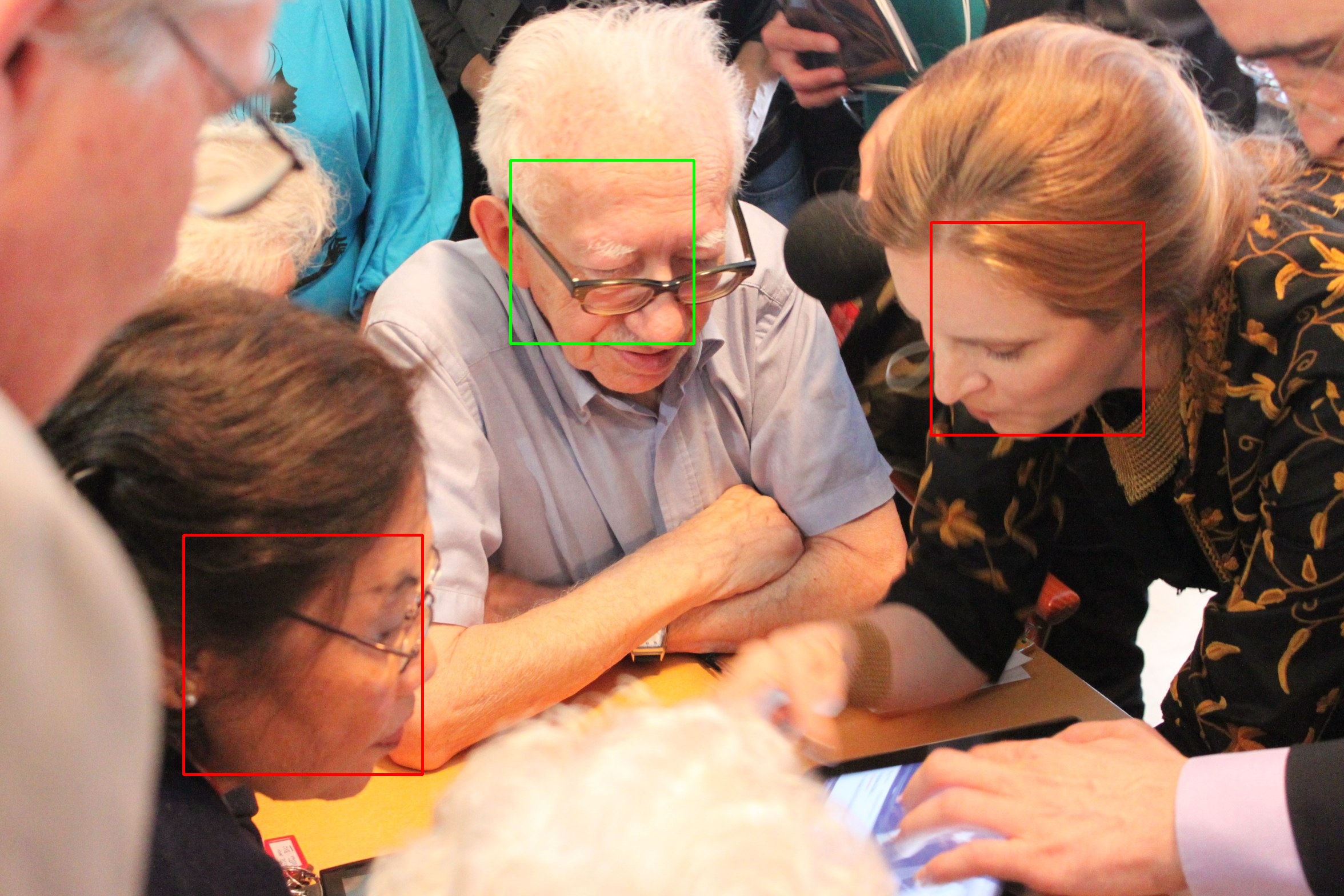}
    \includegraphics[width=0.32\linewidth]{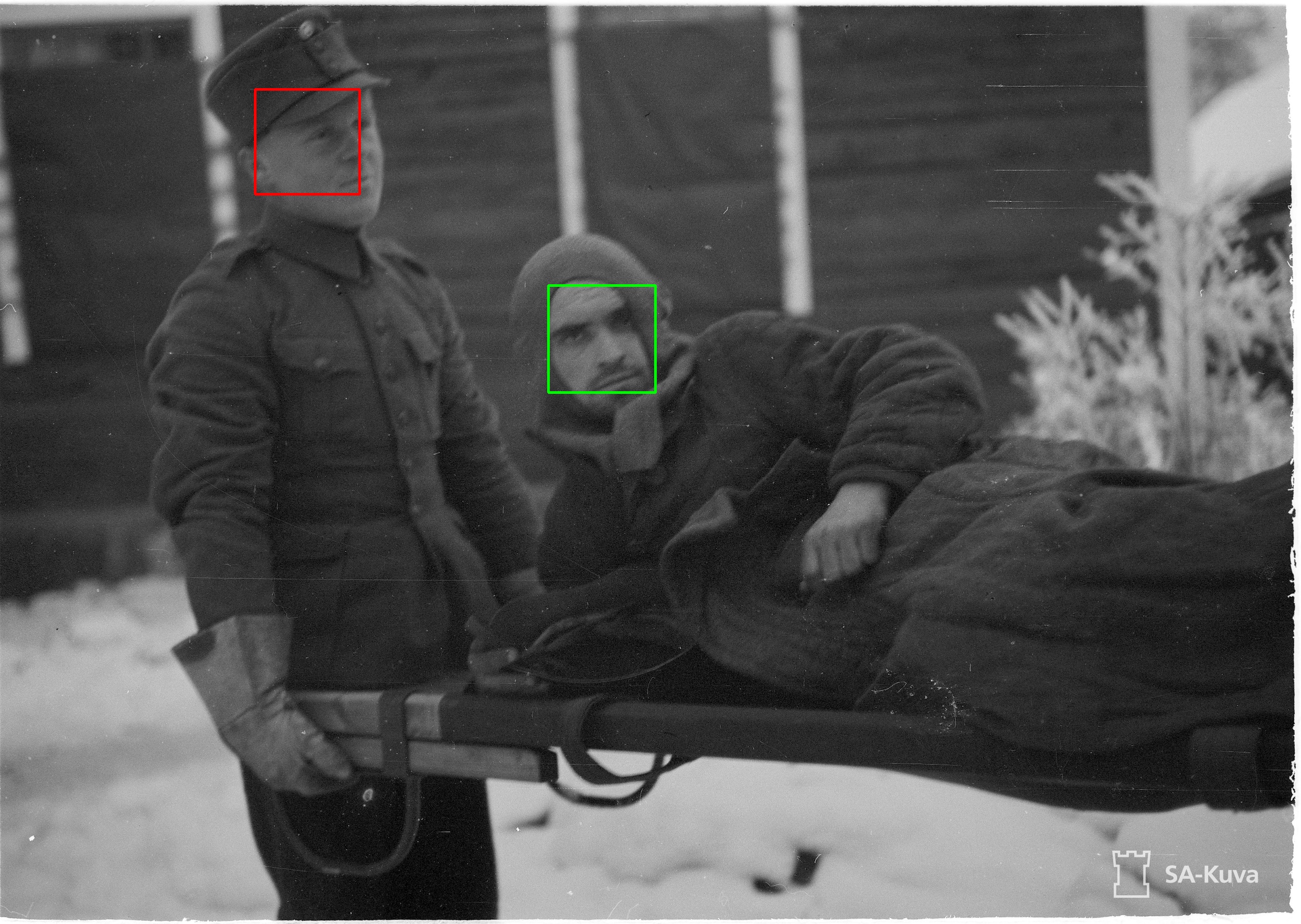}
    \caption{Example outputs of our method on the test datasets. Facial rectangles in green and red denote the predicted main characters and side characters, respectively. \newline Copyright: Helsingin Sanomat, Aamulehti, Google OpenImages and Finnish Wartime Photograph Archive}
        \vspace{-5px}
    \label{fig:goodexamples}
\end{figure*}

\begin{figure}[!t]
    \centering
    \includegraphics[scale=0.3]{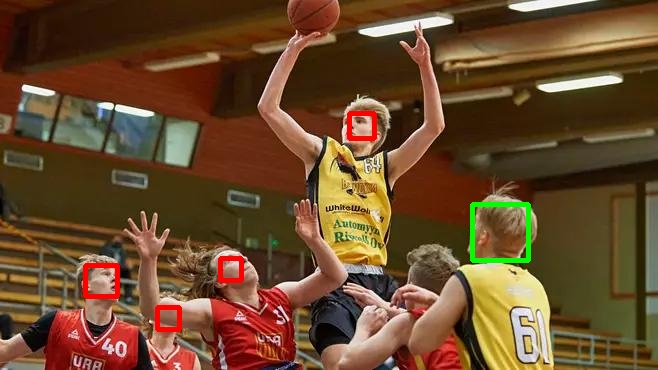}
    \caption{An example photo for which all five annotators agree but the method fails. Only the man with the ball was annotated as the main character by all the annotators but the method labels only the rightmost person as the main character. \newline Copyright: Helsingin Sanomat}
    \label{fig:fail}
        \vspace{-5px}
\end{figure}

Examples of our method performing well on the test dataset can be seen in Fig. \ref{fig:goodexamples}. The predicted main characters have their face rectangles in green, while the side characters have them in red. An example of a case where all our annotators agreed but the method fails is shown in Fig. \ref{fig:fail}. All annotators considered the man with the ball as the only main character, but the method detects only the rightmost person as the main character as shown in the image. The error is primarily caused by the facial area of the rightmost person being significantly larger than the rest. Also, the difference between the sharpness of individual characters is diminished due to the low resolution of this image.

We provide the results of our proposed method and the variant using face detection in Table~\ref{tbl:results}. To analyze the subjectivity level of the main character recognition task, we also compare the annotations from non-expert annotators to the ground-truth annotations from the photographic researcher and give their average scores in Table~\ref{tbl:results}. In each case, we give the results both for all 300 images and only for the photos that the photographic researcher evaluated to be clear cases.
\begin{table}[t]
\centering
\begin{tabular}{|c|c|c|c|}
\hline
\multicolumn{1}{|c|}{Approach} & Precision & \multicolumn{1}{l|}{Recall} & \multicolumn{1}{l|}{F1-score} \\ \hline
Proposed method & \textbf{0.843/0.951} & \textbf{0.820/0.962} & \textbf{0.831/0.956} \\ \hline
{Face detection} & \textbf{0.833/0.872} & \textbf{0.482/0.513}& \textbf{0.611/0.646} \\ \hline
Annotators & \textbf{0.961/0.963} & \textbf{0.984/0.989}& \textbf{0.972/0.976} \\ \hline
\end{tabular}
\caption{Performance of different approaches on all 300 images/on clear cases using the annotations of the photographic researcher as the ground-truth. 'Face detection' refers to the variant of the our method using face detection instead of human pose estimation and 'Annotators' to the average scores of the non-expert annotators.}
\label{tbl:results}
\end{table}

We can see that the performance of the proposed method based on human pose estimation is indeed much better than the variant based on direct face detection. This justifies our methodological choice as expected. We also see that the agreement between the expert annotator and the other annotators is high. This indicates that despite the unclear cases the task does not involve major subjectivity. Interestingly, the proposed method achieves performance close to non-expert humans on clear cases, but clearly lower performance when the full dataset including 70 unclear cases is used. This can be seen as a positive outcome considering the intended use in photographic studies. 

We further analyze our results with respect to the number of annotators who labeled different persons as main characters. We present the ratio of persons labeled as main character by our method separately for each number of annotators seeing the person as a main character in Table \ref{tbl:annotatoragreement3}. The results show that our method's predictions correlate well with the human intuition measured by the number of positive labels.

\begin{table}[t]
\centering
\begin{tabular}{|c|c|c|}
\hline
\begin{tabular}[c]{@{}c@{}}Number of annotators\\  that annotated the person \\ as 'main character'\end{tabular} & \begin{tabular}[c]{@{}c@{}}Number \\ of \\ subjects\end{tabular} & \begin{tabular}[c]{@{}c@{}}Ratio of main \\ character predictions \\ to  total number \\ of subjects\end{tabular} \\ \hline
5 & 377 & 0.8169 \\ \hline
4 & 25 & 0.36 \\ \hline
3 & 15 & 0.2 \\ \hline
2 & 9 & 0.33 \\ \hline
1 & 4 & 0 \\ \hline
0 & 715 & 0.045 \\ \hline
\end{tabular}
\caption{Ratio of main character predictions produced by our method to the total number of people among persons annotated as main characters by a different number of annotators.}
\label{tbl:annotatoragreement3}
\end{table}

\section{Conclusion}
\label{sec:conclusion}

We proposed an algorithm for recognizing the main characters in photos, which is an important step in photographic studies and media analysis.  While some methods have been earlier proposed for ranking people based on their importance in the photos, to the best of our knowledge, automatic main character recognition has not been previously studied from the photographic research perspective. The proposed method relies on state-of-the-art human pose estimation for person detection, but labels the main characters using static features selected based on the main character definition commonly used in photographic studies. This is important because there are no datasets available that match our objectives. Furthermore, the approach makes the results easier to interpret and the weights can be adjusted by photographic researchers for varying needs. We obtained promising results and, in particular on images that were named clear and important cases by a photographic researcher, our method obtained performance comparable to non-expert human annotators. In the future, we plan to further improve the method by changing the score weights adaptively for each image depending on the context (e.g., modern or historical).

\bibliographystyle{IEEEtran}
\bibliography{refs}

\end{document}